\documentclass[11pt]{article}
\usepackage[margin=2.8cm]{geometry}
\usepackage{graphicx}
\usepackage{picinpar}
\usepackage{subcaption}
\usepackage{colortbl}
\usepackage{booktabs}
\usepackage{multirow}
\usepackage{soul}
\usepackage{enumerate}
\usepackage{color}
\usepackage{hyperref}
\usepackage{amsmath}
\usepackage{amssymb}
\usepackage{algorithm}
\usepackage{algpseudocode}
\usepackage{kpfonts}
\usepackage{natbib}

\usepackage[]{authblk}
\usepackage{fancyhdr}
\fancypagestyle{firstpage}{\fancyhead[L]{Preprint}}

\setcitestyle{round, elide, numbers}
\usepackage[]{lineno}

\newcommand{\furl}[1]{\footnote{\url{http://#1}}}

\providecommand{\keywords}[1]
{
  \small	
  \textbf{\textit{Keywords---}} #1
}

\title{Does the Magic of BERT Apply to Medical Code Assignment? A Quantitative Study}

\author[ ]{Shaoxiong Ji}
\author[ ]{Matti H\"{o}ltt\"{a}}
\author[ ]{Pekka Marttinen}
\affil[ ]{Department of Computer Science, Aalto University, Espoo 00076, Finland }
\affil[ ]{\{shaoxiong.ji; matti.m.holtta; pekka.marttinen\}@aalto.fi} 

\date{}

\begin{document}

\maketitle

\begin{abstract}
Unsupervised pretraining is an integral part of many natural language processing systems, and transfer learning with language models has achieved remarkable results in downstream tasks.
In the clinical application of medical code assignment, diagnosis and procedure codes are inferred from lengthy clinical notes such as hospital discharge summaries. 
However, it is not clear if pretrained models are useful for medical code prediction without further architecture engineering.
This paper conducts a comprehensive quantitative analysis of various contextualized language models' performances, pretrained in different domains, for medical code assignment from clinical notes.
We propose a hierarchical fine-tuning architecture to capture interactions between distant words and adopt label-wise attention to exploit label information. 
Contrary to current trends, we demonstrate that a carefully trained classical CNN outperforms attention-based models on a MIMIC-III subset with frequent codes.
Our empirical findings suggest directions for building robust medical code assignment models.

\end{abstract}


\section{Introduction}
\label{sec:introduction}

Clinical notes generated by healthcare professionals are parts of electronic health records and provide an essential source for intelligent healthcare applications~\citep{zhang2020time}. 
Medical information management aims to assign standard medical codes to each clinical document for categorization purposes, which requires professional medical knowledge and is usually costly and error-prone~\citep{hsia1988accuracy, stanfill2010systematic}. 
The International Classification of Diseases (ICD) system, as the most used coding system, provides a global standard for reporting diseases and health conditions.
The rapid development of machine learning and natural language processing (NLP) can replace manual code assignment with automatic coding systems~\citep{farkas2008automatic, crammer2007automatic, farzandipour2010effective}.
Practical medical code assignment requires to capture semantic concepts~\citep{falis2019ontological} and tackle the challenges of lengthy note encoding and large-dimensional code schemes. 

\begin{figure}[htbp]
\centering
	\includegraphics[width=0.5\textwidth]{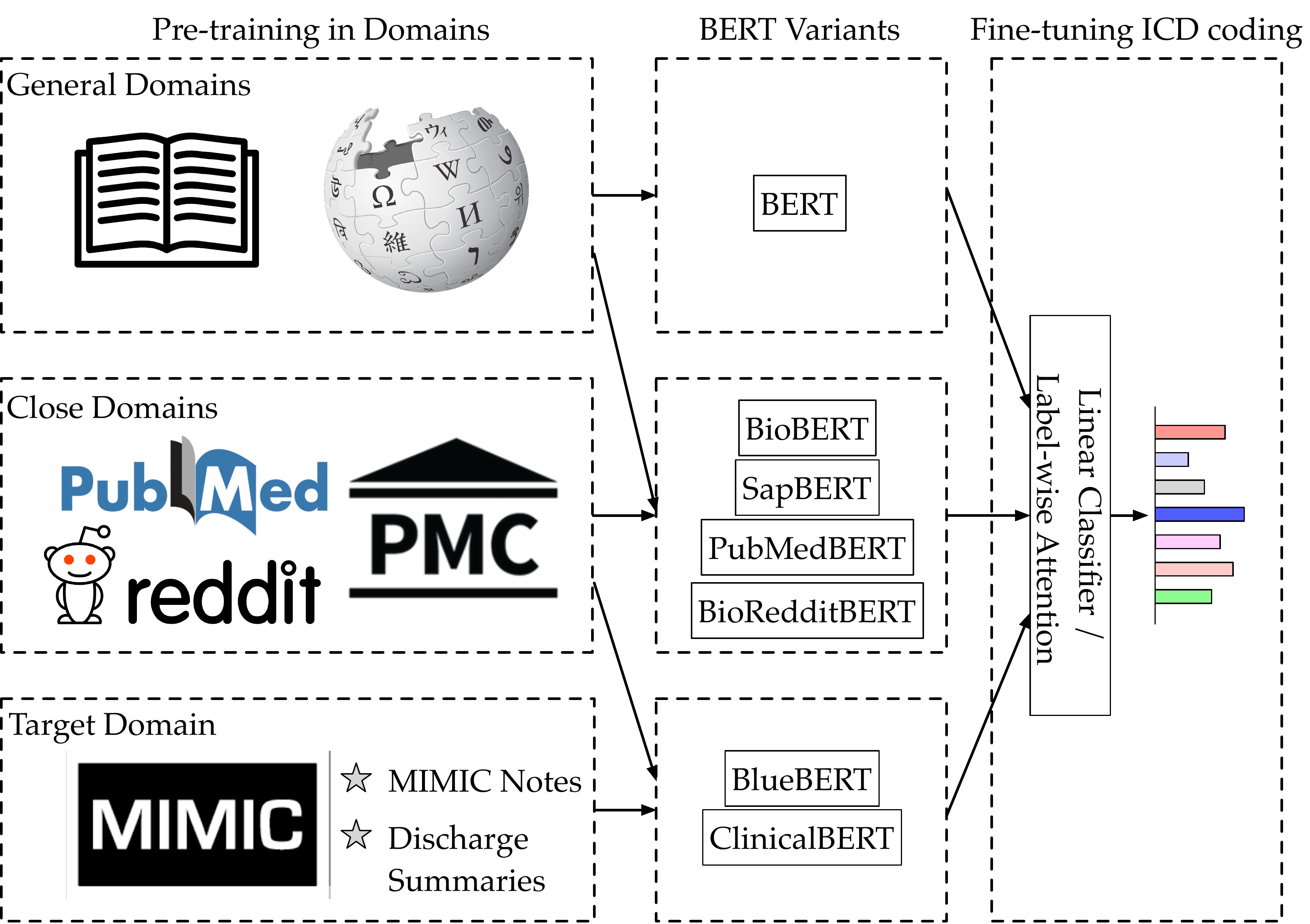}
	\caption{Fine-tuning for ICD coding with pretraining in different domains: general, closely related, target, or mixed-domain pretraining.}
\label{fig:pretraining}
\end{figure}

Pretrained language models (PTM) such as BERT~\citep{devlin2019bert} learn contextualized text representation and have started a new era in NLP.
NLP applications benefit from large-scale pretraining on massive corpora, and universal language representations from PTMs have been successfully utilized in downstream tasks via transfer learning. 
In the field of clinical NLP, incorporating pretrained contextualized language models to encode lengthy clinical notes for large-scale medical code prediction has not been well-studied. 
Recently, \citet{li2020multirescnn}, \citet{ji2021medical} and \citet{dong2020explainable} performed preliminary experiments with pretrained models; however, these three pilot studies failed to achieve satisfactory results or provide in-depth analysis.

This paper investigates language models pretrained in various domains. A language domain here defines a distribution over a topical field such as biomedical documents and clinical notes. Specifically, we investigate the following three research questions (RQ1-3). 
\begin{description}
	\item[RQ1:] \textit{What kind of BERT pretraining works best?} We adopt domain-specific corpora and BERT variants pretrained with different domain adaptation illustrated in Fig.~\ref{fig:pretraining}, and compare their performance on the MIMIC-III benchmark. 
	\item[RQ2:] \textit{What kind of BERT fine-tuning formulation works best for long notes?} We employ classical fine-tuning, develop a hierarchical architecture for long clinical notes, and consider label-aware feature representation. 
	\item[RQ3:] \textit{Are BERT models better than convolution-based approaches?} We reproduce convolutional neural networks (CNNs) with classical pretrained word embeddings and conduct a comparison.
\end{description}

Understanding medical text is a long-lasting research problem. 
We study an essential task in medical information management and diagnosis support - medical code assignment, assigning medical codes to clinical notes. 
Language models can either be pretrained using a large corpus of medical text or trained from scratch and fine-tuned for a specific task. 
Several pretrained models have been published for medical NLP, and they differ in the collections of medical texts used for pretraining, e.g., biomedical, clinical, and medical-related social domains. 
We study the usefulness and relative merits of these different pretrained models and suggest improvements to the neural network architecture to improve performance with long notes: a hierarchical model for longer notes and label-wise attention to leverage relevant information about medical codes. 
Despite our careful attempts, we nevertheless find that fine-tuning pretrained models performs worse than carefully training conventional neural architectures from scratch. 
Hence, our results provide practical guidance for building medical information management systems: pretrained models offer a convenient plug-and-play solution; however, training robust existing standard models offers an appealing practical alternative with good performance in practice.

Our contributions are as follows.
\begin{itemize}
\setlength\itemsep{-0.3em}
\item This paper conducts a comprehensive quantitative study to investigate the effect of knowledge transfer via mixed-domain and task-adaptive language model pretraining in different domains, and a thorough comparative study to answer the research questions.
\item We propose a hierarchical BERT architecture with a label attention mechanism to enhance contextualized representation with label awareness for long clinical notes. 
\item We demonstrate that the classical CNN model with appropriate training can improve the predictive performance, achieving new state-of-the-art results on frequent medical codes (MIMIC-III top-50 dataset).
\end{itemize}

\section{Related Work}
Rule-based and machine learning-based methods have been studied for diagnosis code assignment from clinical notes~\citep{medori2010machine, perotte2014diagnosis}.
\citet{perotte2014diagnosis} proposed an SVM-based classification algorithm with a flat and hierarchy-based classifier. 
Recently, the research trend turns to deep neural networks. 
Convolutional neural networks are one popular category with many model architecture proposed, including CAML that applies CNNs and a label-wise attention mechanism~\citep{mullenbach2018explainable}, MultiResCNN that uses residual connection ~\citep{li2020multirescnn} and DCAN that utilizes dilated convolutions~\citep{ji2020dilated}.
Recurrent neural networks are also extensively studied to capture sequential dependency in clinical notes. 
Such recurrent models include AttentiveLSTM~\citep{shi2017towards}, HA-GRU~\citep{baumel2018multi} and tree-of-sequences LSTM network~\citep{xie2018neural} 
Attention mechanism for matching important diagnosis snippets is widely integrated into CNN- and RNN-based models~\citep{shi2017towards, dong2020explainable}.
CAML~\citep{mullenbach2018explainable} introduced a label-wise attention mechanism to learn label-aware document representations. 

Understanding clinical notes require professional medical knowledge. 
Many methods incorporate external knowledge sources to enhance neural architectures and facilitate clinical text understanding.
\citet{prakash2017condensed} proposed a condensed memory network model with iterative condensation of external memory for network updating and data retrieval from Wikipedia. 
\citet{bai2019improving} used Wikipedia articles of medical codes to learn knowledge-aware embeddings jointly, and \citet{cao2020hypercore} utilized ICD code hierarchy with hyperbolic representation. 
Another direction to incorporate knowledge is through language pretraining and transfer learning.
\citet{li2020multirescnn} and \citet{ji2021medical} reported  preliminary results with semantic knowledge transferring. 
This paper conducts a comprehensive quantitative analysis.

Pretrained language models are trained on an auxiliary task, such as masked language modeling that predicts a word or sequence based on the surrounding context and gains improvement in many NLP tasks~\citep{qiu2020pre}. 
Pretraining such auxiliary tasks benefit from large-scale training on unlabeled corpora that are readily available from the web or textbooks. 
\citet{erhan2010does} hypothesized that pretraining acts as a type of regularization and found that pretrained models exhibit lower generalization errors on average.
Several pretraining models in specific domains have been released, such as BioBERT~\citep{lee2020biobert} and ClinicalBERT~\citep{alsentzer2019publicly}.
They have also been applied in many domain applications; for example,~\citet{mulyar2019phenotyping} applied contextualized language models for phenotyping and
\citet{huang2019clinicalbert} used pretrained models to encode clinical notes to predict hospital readmission.

\section{Method}
\label{sec:method}

we develop fine-tuning with different architectures, including a fully-connected classifier and a hierarchical classifier with an extra transformer atop (Fig.~\ref{fig:HiBERT}) to address the long-document challenge.
Label-wise attention to learn label-aware document representations with these two fine-tuning architectures is described in Section \ref{sec:lan}.

\subsection{Pretraining Domains}
\label{sec:domains}
We study three types of domains: 1) general domains such as book corpora and general Wikipedia articles; 2) domains that are closely related to the target clinical domain; 3) the target clinical domain. 
Assigning ICD codes from clinical notes is a task in the clinical domain. 
We consider biomedical and health-related social domains as candidate domains closely related to the clinical domain.
Inspired by domain- and task-adaptive pretraining~\citep{gururangan2020don}, we investigate different ways of pretraining models for medical code assignment: 1) pretrain only on general domains and immediately transfer to the target clinical domain;
2) continue pretraining on close domains and clinical domains such as the biomedical domain, and transfer to the target clinical domain
3) pretrain on close domains from scratch and transfer to the target clinical domain;
4) pretrain on mixed domains and further fine-tune on the target domain.
The details of methods that fall into these classes can be summarized with three categories: 

\paragraph{Pretraining in General Domains} does not involve specific topics or genres.  
We use BERT~\citep{devlin2019bert} pretrained on two unsupervised prediction tasks, i.e., masked language model and next sentence prediction, using the BooksCorpus~\citep{zhu2015aligning} and English Wikipedia. 

\paragraph{Mixed-domain Pretraining} contains a mixture of domains. We consider continued training and training from scratch. 
Domain-adaptive pretraining has been validated for the ability to improve the predictive performance~\citep{gururangan2020don} further. 
We use 1) BlueBERT~\citep{peng2019transfer} pretrained with PubMed text and MIMIC-III clinical notes; 
2) BioBERT~\citep{lee2020biobert} continually pretrained on domain-specific data from PubMed abstracts and PMC full-text articles;
3) BioRedditBERT~\citep{basaldella2020cometa} initialized from BioBERT and continually pretrained on health-related posts from health-themed forums in Reddit;
4) PubMedBERT~\citep{gu2020domain} domain-specific pretrained from scratch in biomedical domain using PubMed publications; and 
5) SapBERT~\citep{liu2020self} that benefits from self-aligning biomedical entities to the Unified Medical Language System (UMLS) ontology and non-parametric metric learning, leading to a better-separated embedding space.

\paragraph{In-domain Continued Pretraining} continues pretraining in the target clinical domain, which is also called task-specific pretraining~\citep{howard2018universal, gururangan2020don}. 
We use the ClinicalBERT~\citep{alsentzer2019publicly} with the second phase of pretraining in the clinical domain using clinical notes and discharge summaries.

\subsection{Fine-tuning with Hierarchical Structure}
\label{sec:fine-tuning}
Clinical notes, consisting of patient history and discharge summaries, are often long documents. To address this, we develop two fine-tuning strategies, one based on the straightforward truncation and the other with hierarchical architecture. 
In the first approach, we truncate clinical notes to 512 tokens, take the final hidden state of the first token \texttt{[CLS]} as the pooled representation of the truncated note (denoted as $C\in \mathbb{R}^{d_h}$), and apply a fully connected network (FCN) as the classifier with sigmoid activation to predict output probabilities. This straightforward fine-tuning structure is denoted as \texttt{BERT-trun}, and it serves as the baseline.

\begin{figure}
\centering
  	\includegraphics[width=0.49\textwidth]{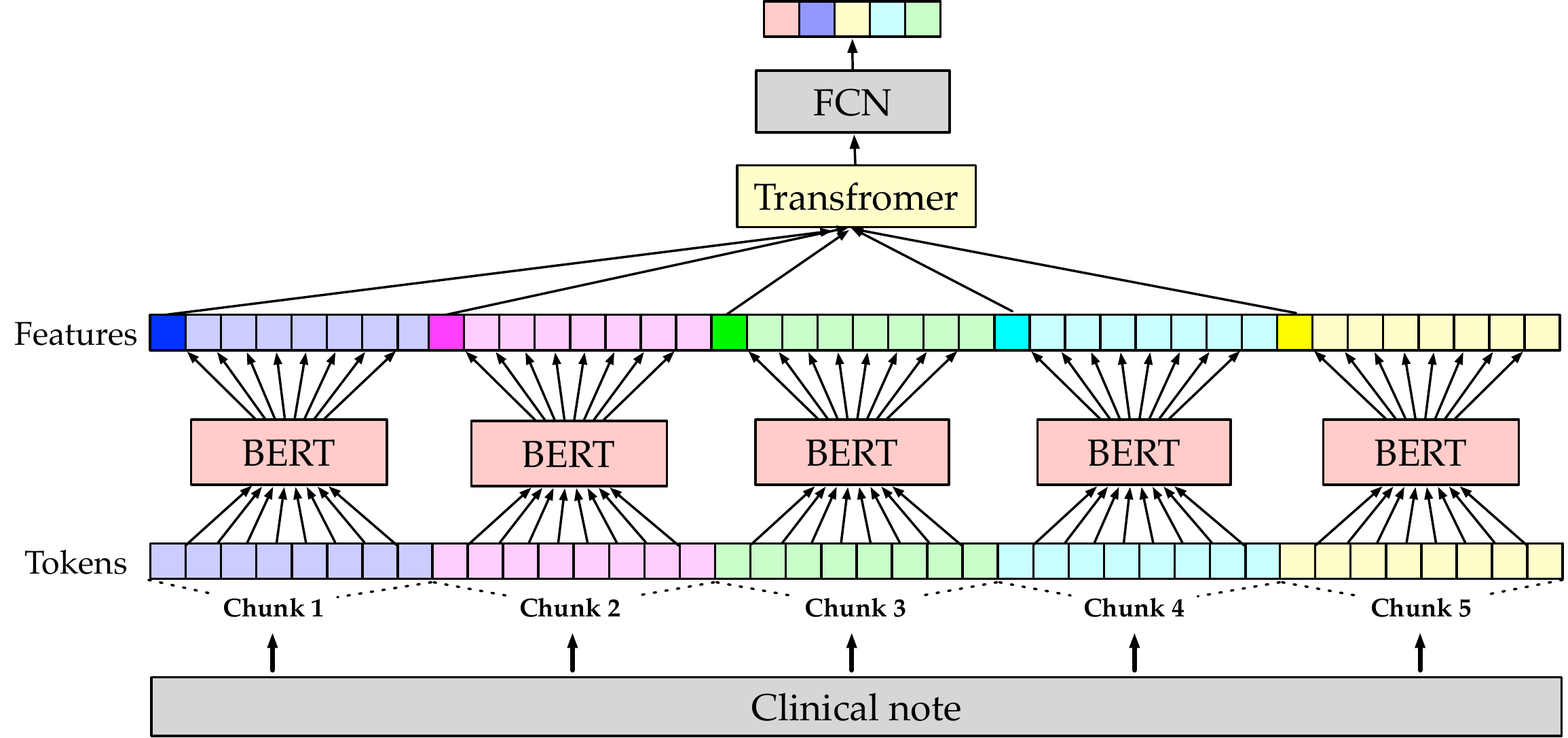} 
\caption{Hierarchical BERT dealing with long clinical notes.}
\label{fig:HiBERT}
\end{figure}

Besides, we propose a hierarchical fine-tuning structure to deal with long notes, shown in Fig.~\ref{fig:HiBERT}.
The lengthy clinical notes (with more than 512 tokens) are first divided into several shorter subsequences to build the lower-level contextualized representation. 
An additional transformer network~\citep{vaswani2017attention} is built atop to capture the second-level sequential dependencies between the note segments. 
The classifier follows the same setup of the truncated version, i.e., FCN with sigmoid. 
We call this hierarchical structure \texttt{BERT-hier}. 

The learning objective function adopts the binary cross entropy loss denoted as:
\begin{equation}
\small
\label{eq:bce}
\mathcal{L}=\sum_{i=1}^{m}\left[-y_{i} \log \left(\hat{y}_{i}\right)-\left(1-y_{i}\right) \log \left(1-\hat{y}_{i}\right)\right],
\end{equation}
where $y_{i} \in\{0,1\}$ is the ground-truth label, $\hat{y}_{i}$ is the sigmoid score for prediction, and $m$ is the number of ICD codes.
We use the AdamW optimizer~\citep{loshchilov2019decoupled} to fine-tune the model with backpropagation. 

\subsection{Label-wise Attention}
\label{sec:lan}
To further connect the document representation with label information, we introduce label-wise attention~\citep{mullenbach2018explainable} in the fine-tuning procedures.
The label attention network (LAN) prioritizes essential information in the hidden note representation relevant to the medical ICD codes. 
The LAN calculates the attention score $\mathbf{A}\in \mathbb{R}^{n\times m}$, which measures the importance between each pair of medical codes and words in the document. It is defined as a dot product:
\begin{equation}
\mathbf{A}=\operatorname{Softmax}(\mathbf{H} \mathbf{U}),
\end{equation}
where $\mathbf{H}\in \mathbb{R}^{n\times h}$ is the hidden encoding of the BERT encoder's last layer, $\mathbf{U}\in \mathbb{R}^{{h}\times m}$ is the parameter matrix of the label attention layer (also known as the query), $n$ is the number of tokens in the document, $h$ is the hidden representation dimension, and $m$ is the number of ICD codes. 
The attention layer's output is then calculated by multiplying attention $\mathbf{A}$ with the hidden representation from the last layer of BERT encoder.
The attentive representation $\mathbf{V}\in \mathbb{R}^{m\times {h}}$ formalized as 
$\mathbf{V}=\mathbf{A}^{\mathrm{T}} \mathbf{H}$ is further used for medical code classification, representing sequential dependency and label awareness. 
The \texttt{BERT-trun} and \texttt{BERT-hier} fine-tuning architectures can both integrate the label-wise attention mechanism. 

\subsection{Experimental Setup}
\label{sec:setup}
This paper focuses on assigning ICD codes to textual discharge summaries from a hospital stay. 
The Medical Information Mart for Intensive Care III (MIMIC-III) repository~\citep{johnson2016mimic} contains patients admitted to the Intensive Care Unit (ICU) at a US medical center from 2001 to 2012. 
The MIMIC database is available by request via this website \url{https://mimic.physionet.org}.
We use the ``noteevents" table in the latest release of version 1.4 with a total of 58,576 hospital admissions. Specifically, free-text discharge summaries are extracted. 
Two settings are used for experimental evaluation. One uses the top 50 frequent labels derived from~\citet{shi2017towards}, while another uses the full set of ICD codes.
Table~\ref{tab:datasets} shows a statistical summary of two sets of MIMIC data. 

\begin{table}[htp]
\small
\caption{A statistical summary of datasets}
\begin{center}
\begin{tabular}{l|c|c|c|c}
\toprule
\text { Dataset } & \text {Train } & \text {Dev. } & \text{Test} & \text{Labels}\\  
\midrule
MIMIC-III top-50 & 8,066 & 1,573 & 1,729 & 50 \\
MIMIC-III full & 47,723 & 1,631 & 3,372 & 8,921 \\
\bottomrule
\end{tabular}
\end{center}
\label{tab:datasets}
\end{table}%

\paragraph{Preprocessing.}
We use raw notes, ICD diagnoses, and procedures for patients from the public clinical MIMIC-III dataset for experiments. 
Discharge summaries labeled with a set of ICD-9 diagnosis and procedure codes include descriptions of procedures performed by a physician, diagnosis notes, patient's medical history, and discharge instructions.
Addenda of admissions are concatenated to a single document. 
The NLTK package is utilized for tokenization, and all tokens are converted into lowercase.
Non-alphabetic characters, such as numbers and punctuations, are removed. 
All documents are truncated at 512 tokens and 2500 tokens for a single BERT encoder (\texttt{BERT-trun}) and hierarchical BERT encoders (\texttt{BERT-hier}).

\paragraph{Training.}
For the prior publications and the recommendation of fine-tuning the BERT model, we choose some common settings. For example, the dropout probability is 0.1. 
The Adam optimizer~\citep{kingma2014adam} is used to optimize CNN-based models, and the AdamW optimizer~\citep{loshchilov2019decoupled} to fine-tune BERT variants. 
We utilize a linear learning rate scheduler with warmup and layer-wise learning rates when fine-tuning BERT-based models.
For retraining the CNN-based models, we use the CBOW of word2vec~\citep{mikolov2013distributed} and adopt static word embeddings. 
We set the batch size for MIMIC-III top-50 and full sets at 8, and the learning rate from $1\mathrm{e}^{-6}$ to $1\mathrm{e}^{-3}$.

All the models are run on a Linux cluster with Nvidia P100 or V100 GPUs. 
For the MIMIC-III top-50 data set, the fine-tuning architecture with truncated notes and a linear classifier has about 109M parameters. 
In comparison, the MIMIC-III full set takes about 115M parameters. 
For the hierarchical architecture, MIMIC-III top-50 and full codes data sets have 115M and 128M parameters, respectively. 
Generally speaking, fine-tuning pretrained models consumes the memory of a large GPU; however, it is arguably less expensive than training from scratch. 

\begin{table*}[htbp]
\small
\setlength{\tabcolsep}{2pt}
\caption{Results of PLM fine-tuning with \texttt{BERT-hier} + LAN in various domains on MIMIC-III dataset with top-50 and full ICD codes. Clinical notes are truncated at length of 2500. Hierarchical fine-tuning structure with label-wise attention is used. \textbf{Bold} text denotes the best and \textit{italic} text denotes the second best.}
\begin{center}
\begin{tabular}{lrr|rr|r|rr|rr|rr}
\toprule
 \multirow{3}{4em}{Model} 	& \multicolumn{5}{c}{MIMIC-III Top-50 Codes} & \multicolumn{6}{c}{MIMIC-III Full Codes} \\
	\cline{2-12} 
	& \multicolumn{2}{c}{AUC-ROC} & \multicolumn{2}{c}{ F1 } & \multirow{2}{2em}{P@5} & \multicolumn{2}{c}{AUC-ROC} & \multicolumn{2}{c}{ F1 } & \multirow{2}{2em}{P@8} & \multirow{2}{2em}{P@15}\\  
 	&Macro &Micro& Macro&Micro &  &Macro &Micro& Macro&Micro &\\
\midrule
BERT-base	&	82.7	&	86.3	&	40.8	&	50.8	&	52.2	&	82.2	&	96.6	&	5.8	&	44.1	&	63.3	&	48.1	\\ 
\hline
BlueBERT	&	\textbf{89.4}	&	\textbf{92.0}	&	61.0	&	65.6	&	62.8	&	84.4	&	97.5	&	5.1	&	42.5	&	62.6	&	47.3	\\
BioBERT full text	&	88.8	&	\textit{91.7}	&	60.4	&	66.0	&	\textit{63.1}	&	85.2	&	97.4	&	\textbf{6.4}	&	\textbf{47.0}	&	\textit{65.8}	&	\textit{50.7}	\\
BioRedditBERT	&	87.1	&	89.6	&	59.4	&	64.8	&	62.4	&	\textit{86.5}	&	\textit{98.0}	&	3.0	&	40.6	&	62.4	&	47.8	\\
PubMedBERT full text	&	88.6	&	90.8	&	\textbf{63.3}	&	\textbf{68.1}	&	\textbf{64.4}	&	\textbf{87.4}	&	\textbf{98.1}	&	4.3	&	44.5	&	65.2	&	50.4	\\
SapBERT	full text &	88.5	&	90.8	&	\textit{62.2}	&	\textit{66.7}	&	\textit{63.1}	&	86.4	&	97.7	&	\textit{6.2}	&	\textit{46.8}	&	\textbf{68.5}	&	\textbf{53.0}	\\
ClinicalBERT all notes	&	\textit{89.2}	&	91.6	&	59.5	&	64.8	&	62.0	&	84.7	&	97.4	&	6.0	&	46.6	&	65.1	&	49.9	\\
\bottomrule
\end{tabular}
\end{center}
\label{tab:m3-BERT-hier}
\end{table*}%

\section{Results}
\label{sec:experiments}
We conduct a series of experiments with different pretrained models on various domains using two fine-tuning architectures and reproduce classical CNN-based models with word embeddings from scratch. 
We make our code publicly available at \url{https://agit.ai/jsx/MCA_BERT}. 

Different evaluation metrics are utilized for experimental evaluation, including micro and macro F1 scores and area under the receiver operating characteristic curve (AUC-ROC).
We evaluate the metrics of precision at $k$, where $k=5$ for MIMIC-III subset with top-$50$ frequent codes and $k=8, 15$ for full sets of MIMIC-III, given the observation that most medical documents are assigned no more than 20 codes.
 For example, the macro precision is calculated as the overall average precision across all labels, given by:
 \begin{equation*}
 \small
 \text { Macro-Precision }=\frac{1}{m} \sum_{\ell=1}^{m} \frac{\mathrm{TP}_{\ell}}{\mathrm{TP}_{\ell}+\mathrm{FP}_{\ell}}, 
 \end{equation*}
 where $\mathrm{TP}_\ell$ and $\mathrm{FP}_\ell$ are the numbers of true and false positives of code $\ell$. 
Micro scores give more weight to frequent labels by considering all labels jointly. 
 For example, the micro precision is defined as:
 \begin{equation*}
 \small
 \text { Micro-Precision }=\frac{\sum_{\ell=1}^{m} \mathrm{TP}_{\ell}}{\sum_{\ell=1}^m \mathrm{TP}_{\ell}+\mathrm{FP}_{\ell}}.
 \end{equation*}

\subsection{Pretraining in Close Domains Improves Prediction (RQ1)}

\begin{table*}[htbp]
\small
\setlength{\tabcolsep}{2pt}
\caption{Effect of more pretraining data with \texttt{BERT-hier} + LAN on MIMIC-III dataset with top-50 and full ICD codes. Clinical notes are truncated at length of 2500. \textbf{Bold} text denotes better performance.}
\begin{center}
\begin{tabular}{lrr|rr|r|rr|rr|rr}
\toprule
 \multirow{3}{4em}{Model} 	& \multicolumn{5}{c}{MIMIC-III Top-50 Codes} & \multicolumn{6}{c}{MIMIC-III Full Codes} \\
	\cline{2-12} 
	& \multicolumn{2}{c}{AUC-ROC} & \multicolumn{2}{c}{ F1 } & \multirow{2}{2em}{P@5} & \multicolumn{2}{c}{AUC-ROC} & \multicolumn{2}{c}{ F1 } & \multirow{2}{2em}{P@8} & \multirow{2}{2em}{P@15}\\  
 	&Macro &Micro& Macro&Micro &  &Macro &Micro& Macro&Micro &\\
\midrule
PubMedBERT abstract	&	\textbf{90.1}	&	\textbf{92.5}	&	62.3	&	67.0	&	63.8	&	85.2	&	97.1	&	\textbf{5.8}	&	37.5	&	\textbf{66.6}	&	50.4	\\
PubMedBERT full text	&	88.6	&	90.8	&	\textbf{63.3}	&	\textbf{68.1}	&	\textbf{64.0}	&	\textbf{87.4}	&	\textbf{98.1}	&	4.3	&	\textbf{44.5}	&	65.2	&	\textbf{51.7}	\\ 
\hline
BioBERT abstract	&	\textbf{89.4}	&	91.6	&	59.5	&	64.8	&	62.7	&	84.5	&	97.3	&	6.3	&	46.5	&	65.3	&	50.0	\\
BioBERT full text	&	88.8	&	\textbf{91.7}	&	\textbf{60.4}	&	\textbf{66.0}	&	\textbf{63.1}	&	\textbf{85.2}	&	\textbf{97.4}	&	\textbf{6.4}	&	\textbf{47.0}	&	\textbf{65.8}	&	\textbf{50.7}	\\
\hline
ClinicalBERT disch. sum.	&	88.7	&	91.4	&	\textbf{60.6}	&	\textbf{65.9}	&	\textbf{62.6}	&	82.9	&	96.6	&	\textbf{6.2}	&	45.0	&	63.8	&	48.7	\\
ClinicalBERT all notes	&	\textbf{89.2}	&	\textbf{91.6}	&	59.5 	&	64.8	&	62.0	&	\textbf{84.7}	&	\textbf{97.4}	&	6.0	&	\textbf{46.6}	&	\textbf{65.1}	&	\textbf{49.9}	\\
\bottomrule
\end{tabular}
\end{center}
\label{tab:data-BERT-hier}
\end{table*}%

This section studies pretraining in different domains (Sec.~\ref{sec:domains}) to evaluate which pretraining scheme works best for medical coding in the clinical domain.
Results on MIMIC-III top-50 and full code set are shown in Table \ref{tab:m3-BERT-hier}, where hierarchical fine-tuning architecture is used.  
Overall, pretraining in mixed domains improves predictive performance to some extent over the BERT-base pretrained in general domains.
These results show the effectiveness of transfer learning to enhance the learning capacity on downstream tasks. 
Among all pretrained models from various domains,
PubMedBERT pretrained from scratch on biomedical article corpora gains a comparatively better performance.
Specifically, it leads to improvements of 3.9\% and 3.3\% for F1 macro and micro scores on the MIMIC-III top-50 code set. 
One possible explanation is that specific domain pretraining makes downstream classifier concentrate on specified semantic knowledge. 
While for ClinicalBERT with three types of domain, the model's attention may be distracted from relatively broad information. 
However, it still performs better than BERT-base with semantic knowledge only from general domains.

\subsection{Effect of Pretraining with More Data (RQ1)}
Can a larger pool of unsupervised pretraining data lead to performance gain in downstream clinical prediction? 
To answer this question, we conduct experiments on the effect of more pretraining data using three groups of pretrained model.
In the biomedical domain, there are 200K abstracts from PubMed and 270K full-text articles from PubMedCentral.
For the clinical domain, 
The CATEGORY value of the MIMIC-III dataset includes `Discharge Summary', `ECG', `Radiology', and `Echo'. `Discharge summary' indicates that the note is a discharge summary.
Besides, there are also other free-text notes. For example, `Report' indicates a full report, and `Addendum' indicates an additional text added to the previous report.

The results with \texttt{BERT-hier} fine-tuning architecture on two sets of the MIMIC-III dataset are shown in Table  \ref{tab:data-BERT-hier}. 
We also study the effect using truncated short notes with fully connected tuning architecture, with results reported in Table \ref{tab:BERT-trun-50}.
These two tables suggest that pretraining with more data leads to better performance in most cases.

\begin{table}[htbp]
\small
\centering
\setlength{\tabcolsep}{1pt}
\caption{Effect of more pretraining data with \texttt{BERT-trun} on MIMIC-III top-50 code set. Clinical notes truncated at length of 512. \textbf{Bold} text denotes better performance.} 
\begin{center}
\begin{tabular}{lrr|rr|r}
\toprule
 \multirow{2}{4em}{Model} 	& \multicolumn{2}{c}{AUC-ROC} & \multicolumn{2}{c}{ F1 } & \multirow{2}{2em}{P@5} \\  
 	&Macro &Micro& Macro&Micro &\\
\midrule
PubMedBERT abstract	&	\textbf{82.4}	&	\textbf{85.9}	&	47.5	&	54.8	&	54.7	\\
PubMedBERT full text	&	82.1	&	84.4	&	\textbf{52.6}	&	\textbf{57.3}	&	\textbf{55.7}	\\ \hline
BioBERT abstract	&	80.9	&	83.6	&	48.9	&	54.8	&	54.0	\\
BioBERT full text	&	\textbf{81.8}	&	\textbf{84.3}	&	\textbf{50.5}	&	\textbf{55.4}	&	\textbf{54.5}	\\ \hline
ClinicalBERT disch. sum.	&	81.8	&	84.2	&	49.7	&	55.8	&	54.9	\\ 
ClinicalBERT all notes	&	\textbf{82.3}	&	\textbf{85.3}	&	\textbf{50.6}	&	\textbf{56.9}	&	\textbf{55.7}	\\
\bottomrule
\end{tabular}
\end{center}
\label{tab:BERT-trun-50}
\end{table}%

\subsection{Hierarchical Fine-Tuning Improves Prediction (RQ2)}

This section answers to the second research question (RQ2) by comparing the \texttt{BERT-trun} and \texttt{BERT-hier} fine-tuning architectures. 
Table~\ref{tab:trun-vs-hier} shows the results on the MIMIC-III top-50 code set using the two fine-tuning architectures when used either with a fully connected classifier or the label attention network. 
Our results are slightly better than the preliminary results reported by \citet{ji2021medical} with appropriate training tricks. 
Mixed domain pretraining models, such as PubMedBERT and ClinicalBERT, gain increases in evaluation scores.
However, most \texttt{BERT-trun} variants' predictive performance suffers due to the lack of information when long notes are truncated into short ones. 
These results show that the proposed hierarchical fine-tuning architecture effectively utilizes long sequences and boosts performance. The label-wise attention mechanism can further improve the prediction in most cases. 

\subsection{Reproducing CNN Outperforms Advanced Methods on Frequent Codes (RQ3)}
Several recent advances for medical code assignment are CNN-based models. 
This section investigates whether improved training of CNN can improve prediction.
Inspired by performance improvement owing to training tricks, we reproduce the CNN model to check if appropriate training strategies can lead to better scores.

\paragraph{Baselines.}
CNN~\citep{kim2014convolutional} is built on pre-trained word vectors with 1D convolution and max-pooling for text classification. 
CAML~\citep{mullenbach2018explainable} integrates CNNs and a label-wise attention mechanism to learn rich representations. It has a variant called DR-CAML that uses ICD code descriptions to regularize the loss function. 
MultiResCNN~\citep{li2020multirescnn} combines residual learning \citep{he2016deep} and multiple channels concatenation with different convolutional filters, achieving good performance in most settings.
HyperCore~\citep{cao2020hypercore} utilizes hyperbolic embedding and co-graph representation with code hierarchy. It gains slightly better performance than the MultiResCNN.

\paragraph{Reproducing CNNs.}
Table~\ref{tab:m3-CNN} shows the reproduced results compared with baselines.
The hyper-parameters are as follows: the learning rate is 0.003, the filter size is 4, the number of filters is 500, the dropout probability is 0.2, and word2vec embeddings are static.  
Our retrained vanilla CNN gains a better performance compared with two recent advanced MultiResCNN~\citep{li2020multirescnn} and HyperCore~\citep{cao2020hypercore} on the MIMIC-III top-50 code dataset.
Moreover, the P@15 score of the retrained CAML model leads to a significant performance increase in the MIMIC-III full code dataset. 
These results suggest that a simple model with appropriate training could achieve decent performance in NLP applications such as this clinical application.

\begin{table}[htbp]
\small
\centering
\setlength{\tabcolsep}{1pt}
\caption{Results on MIMIC-III top-50 code set fine-tuning with \texttt{BERT-trun} and \texttt{BERT-hier} with FCN or LAN.} 
\begin{center}
\begin{tabular}{lrr|rr|r}
\toprule
 \multirow{2}{4em}{Model} 	& \multicolumn{2}{c}{AUC-ROC} & \multicolumn{2}{c}{ F1 } & \multirow{2}{2em}{P@5} \\  
 	&Macro &Micro& Macro&Micro &\\
\midrule

BERT-base-trun + FCN	&	80.1	&	83.0	&	46.3	&	51.6	&	51.1	\\	
BERT-base-hier + FCN	&	85.4	&	88.4	&	53.1	&	59.4	&	57.6	\\	
BERT-base-hier + LAN	&	83.7	&	86.5	&	48.1	&	54.2	&	53.5	\\	\hline
BlueBERT-trun + FCN	&	78.2	&	82.1	&	23.7	&	36.6	&	45.9	\\	
BlueBERT-hier + FCN	&	89.1	&	91.8	&	59.9	&	65.1	&	62.5	\\	
BlueBERT-hier + LAN	&	\textbf{89.4}	&	\textbf{92.0}	&	61.0	&	65.6	&	62.8	\\	\hline
BioBERT-trun + FCN	&	81.8	&	84.3	&	50.5	&	55.4	&	54.5	\\	
BioBERT-hier + FCN	&	87.3	&	89.5	&	59.9	&	65.6	&	62.7	\\	
BioBERT-hier + LAN	&	88.8	&	91.7	&	60.4	&	66.0	&	63.1	\\	\hline
BioRedditBERT-turn + FCN	&	80.7	&	83.5	&	49.3	&	55.5	&	53.8	\\	
BioRedditBERT-hier + FCN	&	87.0	&	89.4	&	59.1	&	64.2	&	61.9	\\	
BioRedditBERT-hier + LAN	&	87.1	&	89.6	&	59.4	&	64.8	&	62.4	\\	\hline
PubMedBERT-trun + FCN	&	82.1	&	84.4	&	52.6	&	57.3	&	55.7	\\	
PubMedBERT-hier + FCN	&	88.8	&	91.4	&	57.7	&	64.2	&	62.2	\\	
PubMedBERT-hier + LAN	&	88.6	&	90.8	&	\textbf{63.3	}&	\textbf{68.1}	&	\textbf{64.4}	\\	\hline
SapBERT-trun + FCN	&	82.3	&	84.8	&	51.2	&	56.8	&	55.2	\\	
SapBERT-hier + FCN	&	88.5	&	90.3	&	\textbf{63.3}	&	67.3	&	63.8	\\	
SapBERT-hier + LAN	&	88.5	&	90.8	&	62.2	&	66.7	&	63.1	\\	\hline
ClinicalBERT-trun + FCN	&	82.3	&	85.3	&	50.6	&	56.9	&	55.7	\\	
ClinicalBERT-hier + FCN	&	87.2	&	90.7	&	54.6	&	62.5	&	60.7	\\	
ClinicalBERT-hier + LAN	&	89.2	&	91.6	&	59.5	&	64.8	&	62.0	\\	
\bottomrule
\end{tabular}
\end{center}
\label{tab:trun-vs-hier}
\end{table}%

\begin{figure}[htbp]
\begin{center}
\begin{subfigure}[]{0.48\textwidth}
	\includegraphics[width=\textwidth]{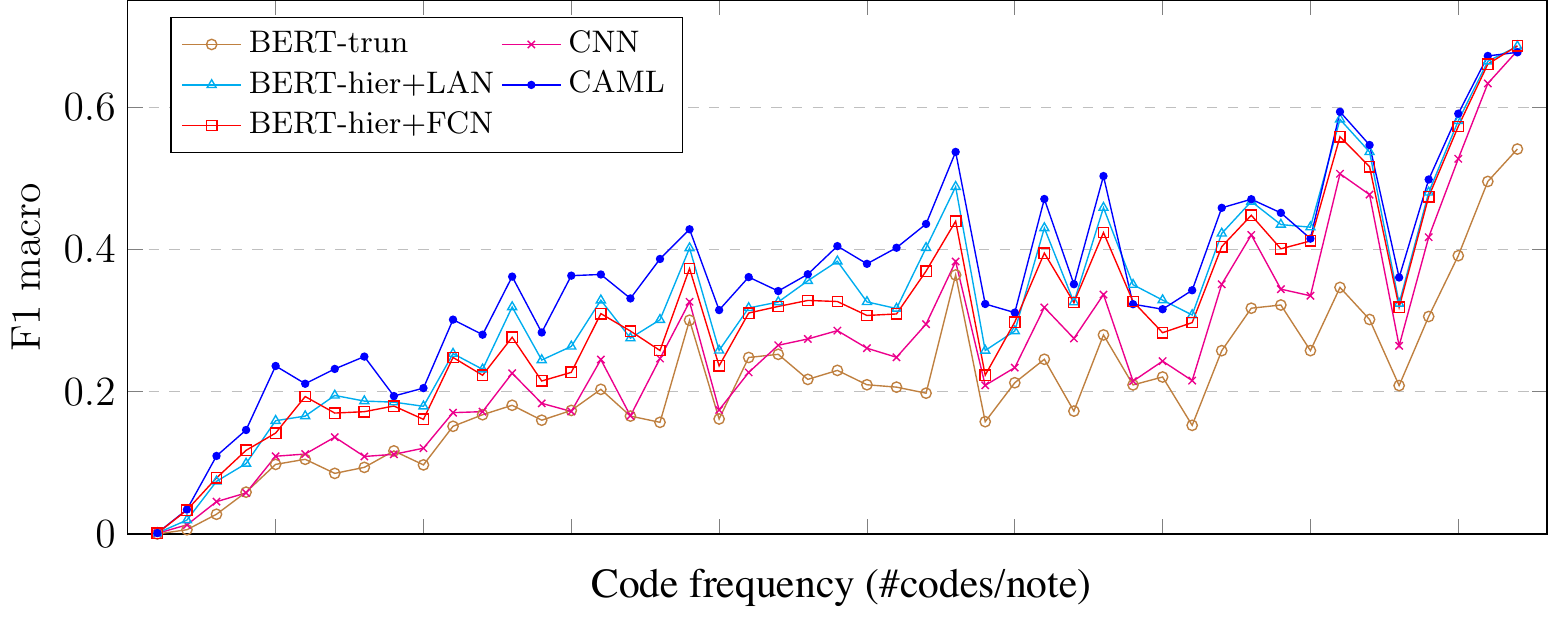}
	\label{fig:f1-macro}
	\subcaption{F1 macro}
\end{subfigure}%
\quad
\begin{subfigure}[]{0.48\textwidth}
	\includegraphics[width=\textwidth]{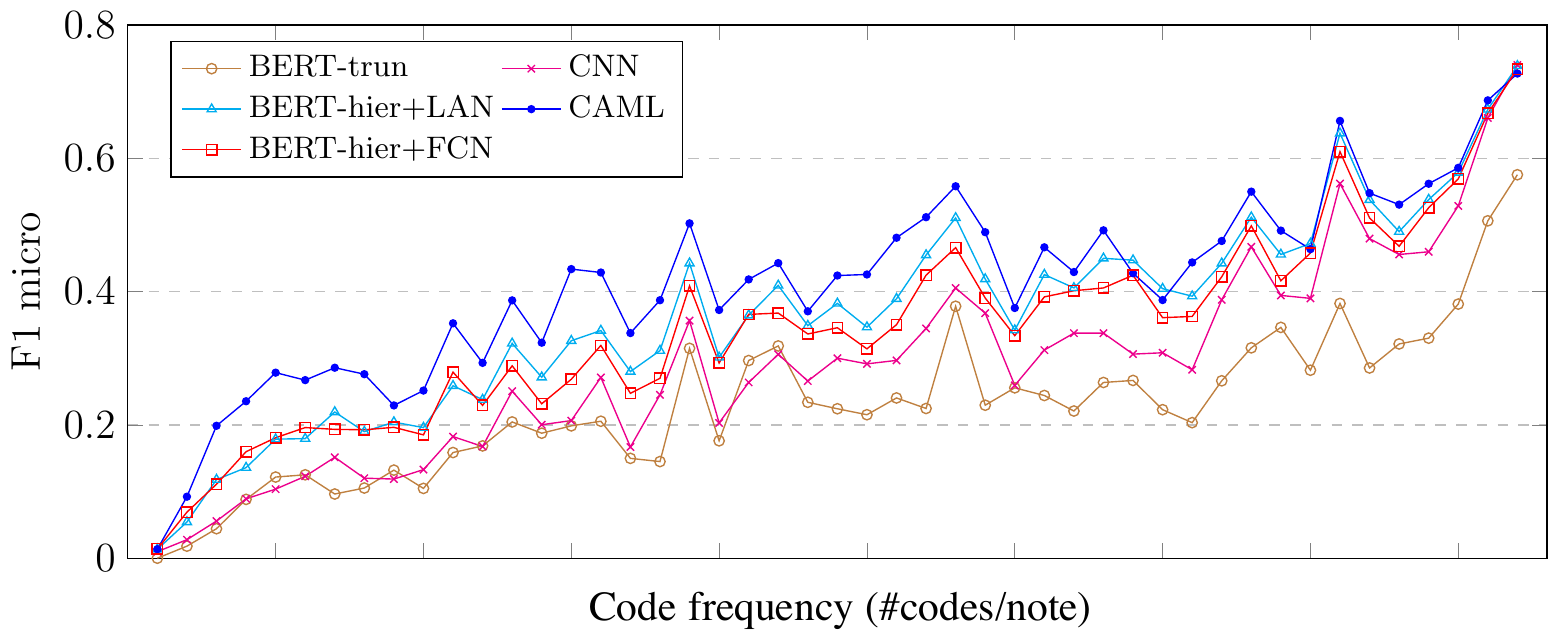}
	\label{fig:f1-micro}
	\subcaption{F1 micro}
\end{subfigure}
\caption{F1 scores of different models on the MIMIC-III full code dataset (8,922 labels). Code frequency groups are sorted in ascending order from left to right.}
\label{fig:f1-scores}
\end{center}
\end{figure}

How did the CNN- and BERT-based models perform on the full code set? 
We bin ICD codes into different groups according to the frequency observed in the training set of the MIMIC-III full code dataset to study the effect of code frequency on the model's predictive performance. 
We take ClinicalBERT as a clinical note encoding representative and compare three fine-tuning strategies with CNN and CAML model. 
Fig. \ref{fig:f1-scores} shows models' predictive behavior on frequent and less frequent codes taking F1 scores as the evaluation metrics. 
These two figures show that all compared models' predictive performance decreases when the ICD code groups have fewer training samples.
When dealing with few-shot codes, model performance drops sharply.
These results again suggest that medical code prediction algorithms should focus on less frequent codes and enhance the robustness of less frequent codes. 

\begin{table*}[htbp]
\small
\centering
\setlength{\tabcolsep}{1pt}
\caption{Results of retrained CNNs on MIMIC-III with top-50 and full ICD codes. \textbf{Bold} text denotes the new state of the art obtained by an old model.}
\begin{center}
\begin{tabular}{lrr|rr|r|rr|rr|rr}
\toprule
 \multirow{3}{4em}{Model} 	& \multicolumn{5}{c}{MIMIC-III Top-50 Codes} & \multicolumn{6}{c}{MIMIC-III Full Codes} \\
	\cline{2-12} 
	& \multicolumn{2}{c}{AUC-ROC} & \multicolumn{2}{c}{ F1 } & \multirow{2}{2em}{P@5} & \multicolumn{2}{c}{AUC-ROC} & \multicolumn{2}{c}{ F1 } & \multirow{2}{2em}{P@8} & \multirow{2}{2em}{P@15}\\  
 	&Macro &Micro& Macro&Micro &  &Macro &Micro& Macro&Micro &\\
\midrule	
CNN~\citep{kim2014convolutional}	&	87.6	&	90.7	&	57.6	&	62.5	&	62.0	&	80.6	&	96.9	&	4.2	&	41.9	&	58.1 & 44.3	\\
CAML~\citep{mullenbach2018explainable}	&	87.5	&	90.9	&	53.2	&	61.4	&	60.9	&	89.5	&	98.6	&	8.8	&	53.9	&	70.9 & 44.5	\\
MultiResCNN~\citep{li2020multirescnn}	&	89.9	&	92.8	&	60.6	&	67.0	&	64.1	& 91.0	&	98.6	&	8.5	&	55.2	&	73.4 & 58.4 \\
HyperCore~\citep{cao2020hypercore}	&	89.5	&	92.9	&	60.9	&	66.3	&	63.2	& 93.0	&	98.9	&	9.0	&	55.1	&	72.2 & 57.9 \\
\hline
CNN (retrained)	&	\textbf{90.8}	&	\textbf{93.1} 	&	\textbf{62.4} 	&	\textbf{67.1} 	&	64.0	&	85.0	&	97.4	&	5.9	&	36.5	&	49.0	&	39.4	\\
\bottomrule
\end{tabular}
\end{center}
\label{tab:m3-CNN}
\end{table*}%

\subsection{Discussion}
Pretrained models in both general and specific domains have shown effectiveness in capturing contextual information. 
However, they encounter limitations in this study. 
In the dataset with top-50 frequent codes, fine-tuning with pretrained models can achieve a good performance; however, the PTM fine-tuning does not work well for high-dimensional structured prediction with a full label set that has more than 8,000 labels. 
This study suggests focusing on less frequent codes. 

Self attention-based models suffer from the complexity of $O(n^2d)$, where $n$ is the sequence length, and $d$ is the dimension of hidden representation, making it hard to encode extremely long documents. 
We investigated how to incorporate the pretrained BERT model and its variants with hierarchical fine-tuning architecture to tackle lengthy clinical document encoding.
Nevertheless, CNN-based models~\citep{mullenbach2018explainable, li2020multirescnn, ji2020dilated} and RNN-based models~\citep{shi2017towards} perform considerably well with the relatively small model scale and remain a meaningful direction. 
Recently, some improved transformer-based models such as Longformer~\citep{beltagy2020longformer}, Linformer~\citep{wang2020linformer} and Big Bird~\citep{zaheer2020big} aim to solve the problem of encoding long document and mitigating the quadratic complexity. 
We leave these emerging models as future work.

\section{Conclusion}
\label{sec:conclusion}
This paper presented a comprehensive quantitive analysis of medical code assignment from clinical notes using various pretrained models with BERT. 
We compared the behavior of several different domain-specific BERT variants. 
To solve the problem of lengthy clinical note encoding, we developed two fine-tuning architectures: 1) fully connected network with simple truncation into short notes; 2) hierarchical fine-tuning architecture with long note segmentation and an additional Transformer on top.
Moreover, we employed label attention to facilitate label-aware representation learning. 
Through intensive experiments, we found that the magic of BERT does not apply to the task of assigning ICD codes from clinical notes.
In contrast, we found that a simple CNN  trained from scratch can achieve superior predictive performance on frequent codes, achieving a new state of the art in the MIMIC-III top-50 dataset. This demonstrates how recent training strategies can improve old models. Our results furthermore suggest that medical code assignment algorithms should pay more attention to less frequent codes.

\section*{Acknowledgments}
This work was supported by the Academy of Finland (grant 336033) and EU H2020 (grant 101016775).
We acknowledge the computational resources provided by the Aalto Science-IT project.
The authors wish to acknowledge CSC - IT Center for Science, Finland, for computational resources.

\bibliography{references}

\end{document}